\pdfoutput=1
\documentclass[11pt]{article}
\usepackage[]{acl}
\usepackage{amssymb}
\usepackage{pifont}
\newcommand{\cmark}{\ding{51}}%
\newcommand{\xmark}{\ding{55}}%
\usepackage{times}
\usepackage{latexsym}
\usepackage{soul}
\usepackage[T1]{fontenc}
\usepackage[utf8]{inputenc}
\usepackage{microtype}
\usepackage{booktabs}
\usepackage{multirow}
\usepackage{pifont}
\usepackage{subcaption}
\usepackage{graphicx}
\usepackage{tabularx}
\usepackage{xcolor}
\usepackage{colortbl}
\usepackage{makecell}
\usepackage{hyperref}
\newcommand{\iw}{Image~Wikification}
\newcommand{\model}{KRAMT}
\newcommand{\data}{\textsc{COFAR}}
\usepackage{enumitem}

\setlist{noitemsep}

\definecolor{orange}{rgb}{1,0.647,0}

\title{COFAR: Commonsense and Factual Reasoning in Image Search}
\author{
  Prajwal Gatti\textsuperscript{\rm 1},\quad
  Abhirama Subramanyam Penamakuri\textsuperscript{\rm 1},\quad
  Revant Teotia\textsuperscript{\rm 2,*},\\
  \textbf{Anand Mishra}\textsuperscript{\rm 1}\textbf{,}\quad
  \textbf{Shubhashis Sengupta}\textsuperscript{\rm 3}\textbf{,}\quad
  \textbf{Roshni Ramnani}\textsuperscript{\rm 3}\\
  \textsuperscript{\rm 1}Indian Institute of Technology Jodhpur, \ \ \textsuperscript{\rm 2}Columbia University,  \ \ \textsuperscript{\rm 3}Accenture Labs\\
  \texttt{\small\{pgatti, penamakuri.1, mishra\}@iitj.ac.in, rt2819@columbia.edu}\\
  \texttt{\small\{shubhashis.sengupta, roshni.r.ramnani\}@accenture.com}\\
}

\begin{document}
\twocolumn[{
\renewcommand\twocolumn[1][]{#1}
\maketitle
\begin{center}
    \centering
    \captionsetup{type=figure}
    \includegraphics[width=0.9\textwidth]{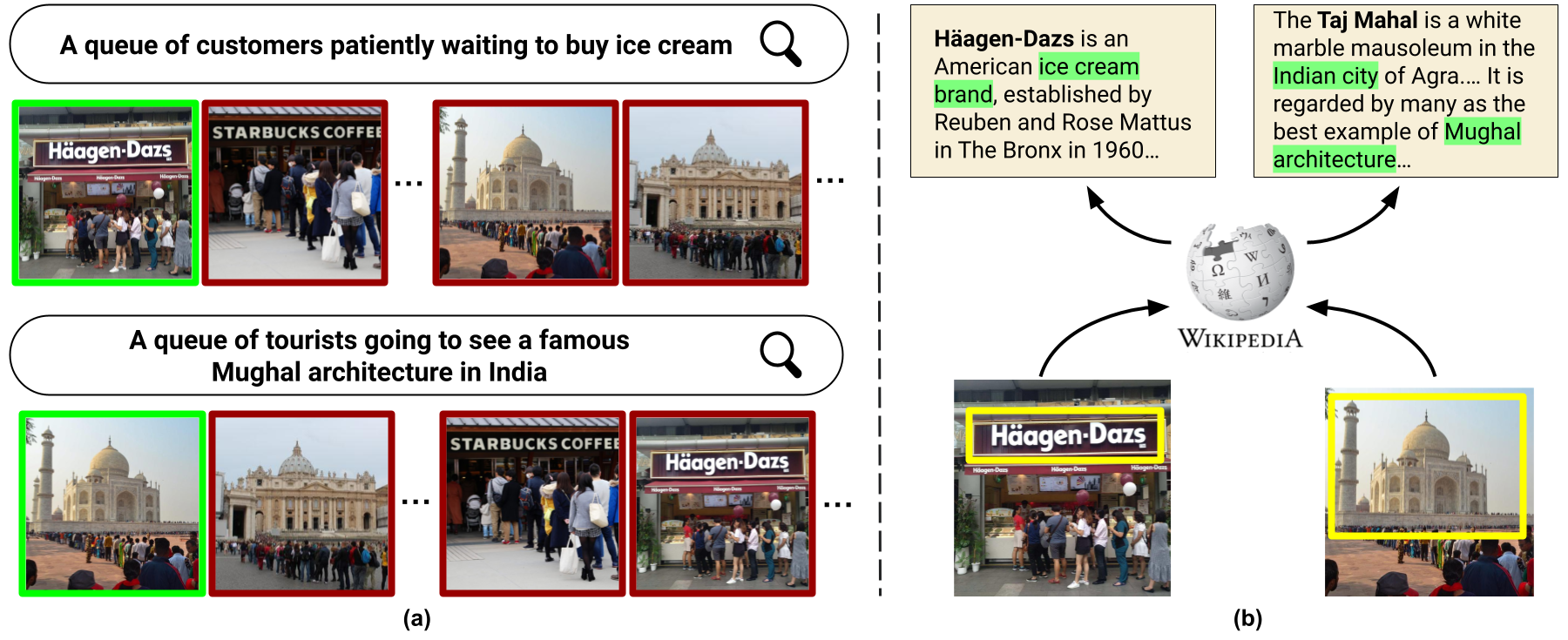}
    \captionof{figure}{\label{fig:goal}Consider the following two natural language queries shown in (a). Retrieving images relevant to these queries (shown using a green bounding box) requires a model that has the ability to interpret images beyond just what is visually apparent, such as interpreting -- who are customers vs. who are tourists? Who are waiting to buy vs. who are going to see? in other words, visual commonsense. Additionally, the model would need to interpret facts or world knowledge, such as Häagen-Dazs is an ice cream brand and the Taj Mahal in India is an example of Mughal architecture. This can be enabled by linking visual entities in the image to an encyclopedic knowledge source such as Wikipedia. Our work presents such a model, namely \model{}.}
\end{center}
}]
\begin{abstract}
\let\thefootnote\relax\footnote{\textsuperscript{*}This work was done while Revant Teotia was affiliated with Indian Institute of Technology Jodhpur.}
One characteristic that makes humans superior to modern artificially intelligent models is the ability to interpret images beyond what is visually apparent. Consider the following two natural language search queries -- (i) ``a queue of customers patiently waiting to buy ice cream" and (ii) ``a queue of tourists going to see a famous Mughal architecture in India." Interpreting these queries requires one to reason with (i) \textbf{Commonsense} such as interpreting people as customers or tourists, actions as waiting to buy or going to see; and (ii) \textbf{Fact} or world knowledge associated with named visual entities, for example, whether the store in the image sells ice cream or whether the landmark in the image is a Mughal architecture located in India. Such reasoning goes beyond just visual recognition. To enable both commonsense and factual reasoning in the image search, we present a unified framework, namely Knowledge Retrieval-Augmented Multimodal Transformer (\model{}), that treats the named visual entities in an image as a gateway to encyclopedic knowledge and leverages them along with natural language query to ground relevant knowledge. Further, \model{} seamlessly integrates visual content and grounded knowledge to learn alignment between images and search queries. This unified framework is then used to perform image search requiring commonsense and factual reasoning. The retrieval performance of \model{} is evaluated and compared with related approaches on a new dataset we introduce – namely \data{}. We make our code and dataset available at \url{https://vl2g.github.io/projects/cofar}.
\end{abstract}

\section{Introduction}
\label{sec:intro}
Retrieving relevant images for a natural language query has been an exciting field of research in the vision-and-language community~\cite{Johnson_2015_CVPR, wang2016learning, wang2020cross}. Most of the available literature focuses on querying visually-evident aspects in the images, such as searching for objects or their interactions in natural scenes. However, as illustrated in Figure~\ref{fig:goal}, users often require an image search engine that can perform commonsense reasoning and leverage facts (world knowledge) about the image content. To fill this gap, we propose a novel image search task requiring commonsense and factual reasoning associated with named visual entities.

To study this problem, a suitable dataset is required. While many text-to-image search datasets are publicly available~\cite{lin2014microsoftCOCO, flickr_30_21014, textCaps}, they have not been explicitly created to study our proposed task. Few of the recently introduced knowledge-enabled VQA datasets such as OK-VQA~\cite{OKVQA2019}, KVQA~\cite{KVQA2019}, text-KVQA~\cite{textKVQA}, FVQA~\cite{FVQA_wang_2017} require either factual or commonsense or a combination of both. However, they may not be well-suited for studying the ``image search" task we are interested in. Note that in the conventional VQA task, a query (question) is evaluated against a single image which is often directly relevant to the query; whereas, in image search, a query needs to be evaluated against several thousands of images, including distractors and then needs to rank the relevant image as the top result. Moreover, to our knowledge, there is no dataset available that includes natural scene images containing a diverse set of visual named entities (such as business brands, celebrities, and world landmarks), visual details of the natural scene along with annotations that demands commonsense and factual reasoning associated with the images. To meet these requirements, we present \data{}, which contains manually annotated English language queries for natural scenes containing named visual entities.

A plausible approach to addressing our image search problem on \data{} is large-scale vision-language pretraining~\cite{radford2021learning, Lu_2020_CVPR} and learning the associations between commonsense-factual concepts and images. This can be successful in learning popular associations, e.g., Starbucks to Coffee, Eiffel tower to Paris if it has seen such samples during training. However, such methods often require large data and generalize poorly to unseen or rare entities. In contrast, we take a distinct path in this work and ground external knowledge associated with entities in the images to perform commonsense and factual reasoning. To this end, we present a unified model, namely \underline{K}nowledge \underline{R}etrieval-\underline{A}ugmented \underline{M}ultimodal \underline{T}ransformer (\model{}), that retrieves relevant knowledge from Wikipedia by performing query-knowledge similarity-guided visual entity linking. It then encodes the retrieved knowledge, query and visual features, and learns image-query alignment using a multimodal transformer to perform knowledge-aware image search.

\noindent\textbf{Contributions of this paper:} (i) We study the problem of image search requiring both commonsense and factual reasoning associated with named visual named entities such as business brands, celebrities, and world landmarks for the first time and introduce a novel dataset, viz. \data{} for this task. We firmly believe that the proposed task, accompanying dataset, and benchmarks presented in this paper will open up future research avenues. (Section~\ref{sec:COFAR})
(ii) We introduce a knowledge retrieval augmented multimodal transformer (\model{}) -- a unified framework that learns to align queries with the relevant images by performing visual entity linking, retrieving relevant knowledge, and seamlessly integrating it with visual content. The experimental results demonstrate that \model{}, besides visual reasoning, can perform commonsense and factual reasoning (Section~\ref{sec:approach} and Section~\ref{sec:expts}).

\begin{figure*}[!t]
\centering
    \begin{subfigure}{.3\textwidth}
        \centering
        \includegraphics[width=0.9\linewidth]{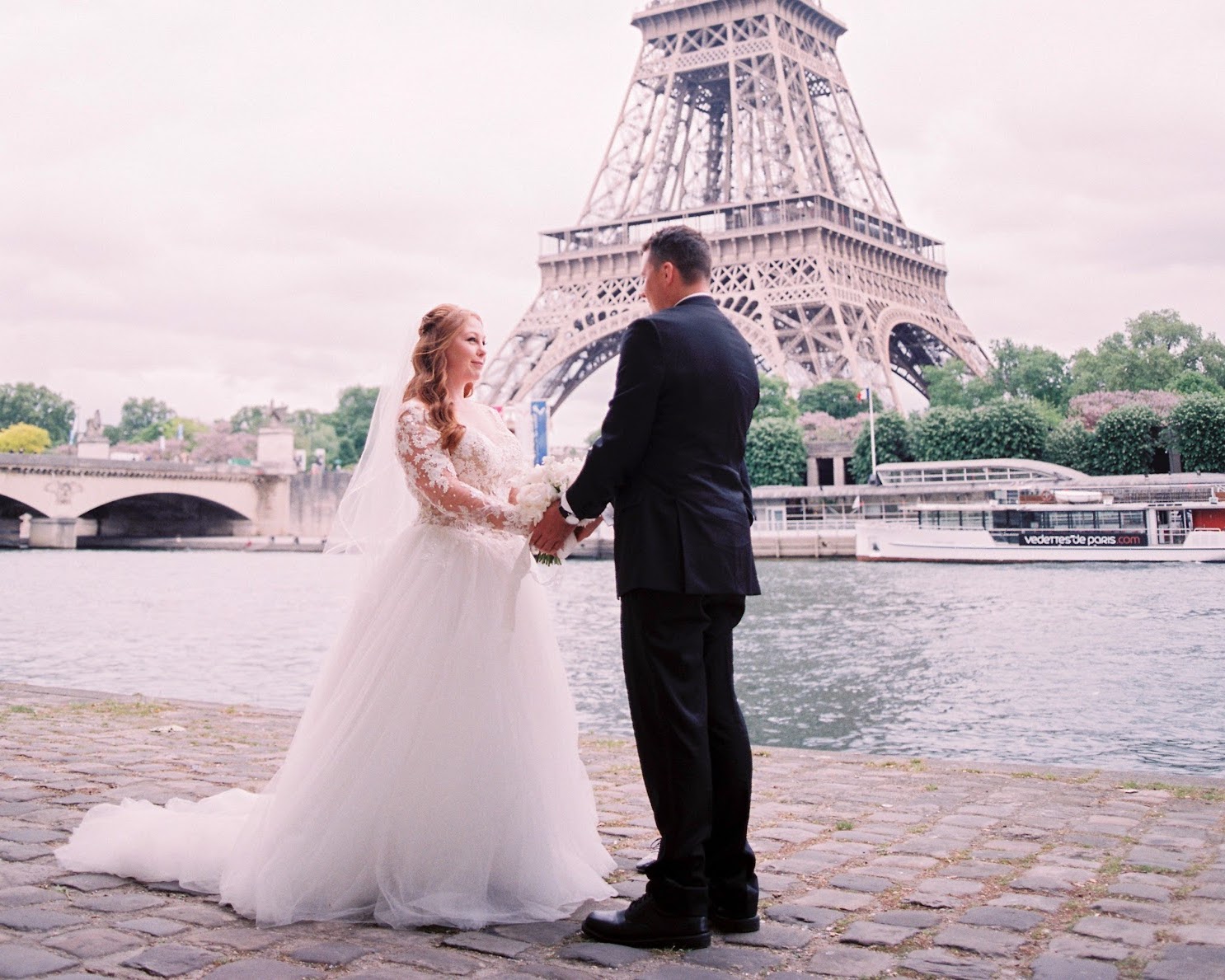} 
        \caption{\tiny{\textit{\textbf{Query}}: Two people getting married in front of a tower in Paris.
            \newline
            \textbf{Commonsense}: Two people in white gown and suit holding hands leads to the commonsense that they are getting married.
            \newline
            \textbf{Visual named entity:} The Eiffel Tower
            \newline
            \textbf{Fact}: The landmark is Eiffel Tower, which is located in Paris, France.
        }}
        \label{fig:cofar-second}
    \end{subfigure}
\hfill
    \begin{subfigure}{.3\textwidth}
        \centering
        \includegraphics[width=0.9\linewidth]{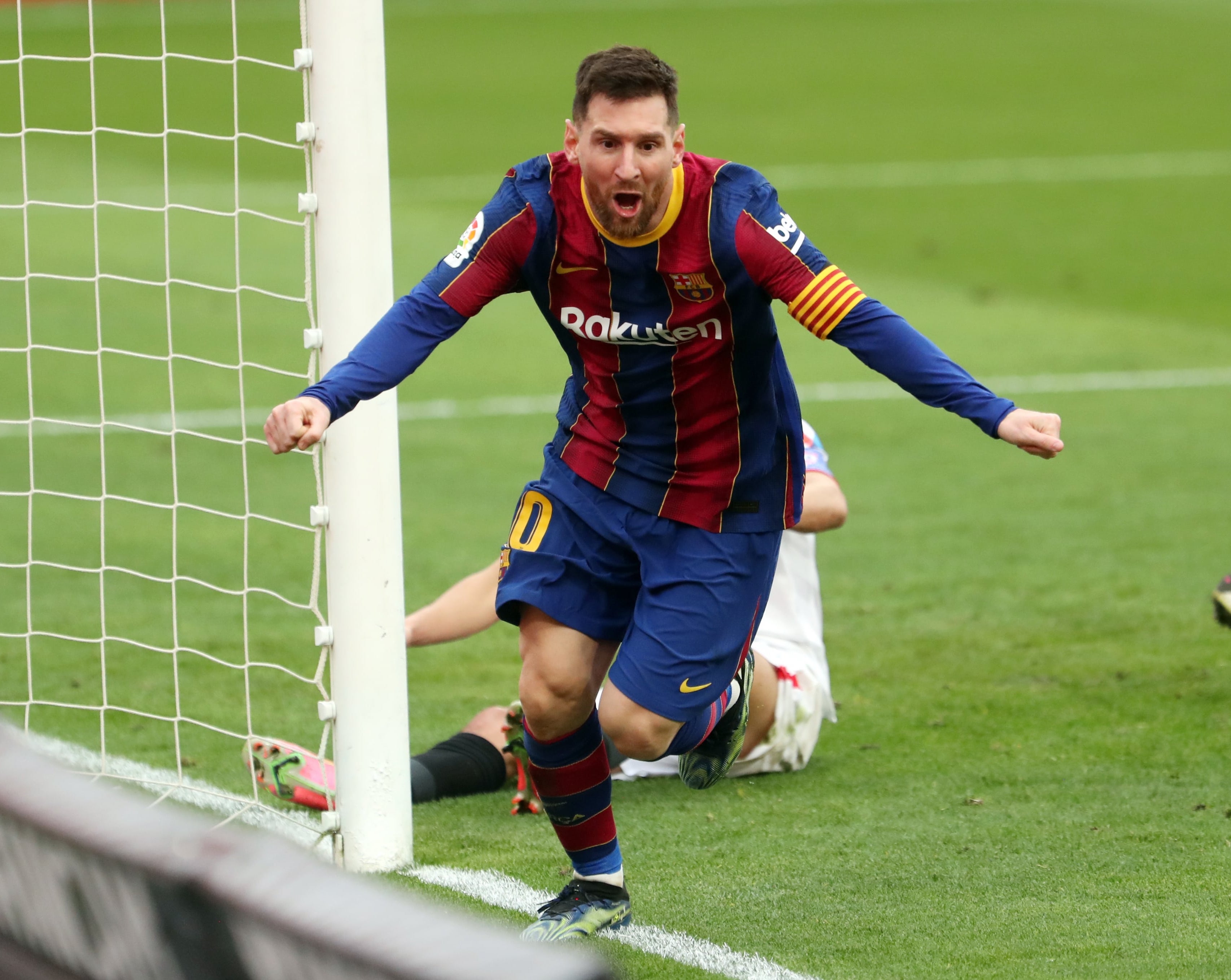} 
        \caption{\tiny{\textit{\textbf{Query}}: The captain of the Argentina national football team celebrating after scoring a goal.
            \newline
            \textbf{Commonsense}: The person is running cheerfully next to a goalpost leads to commonsense that they are celebrating after scoring a goal.
            \newline
             \textbf{Visual named entity:} Lionel Messi}
            \newline
            \textbf{Fact}: Lionel Messi is the captain of the Argentina national football team.
        }
        \label{fig:cofar-third}
    \end{subfigure}
\hfill
    \begin{subfigure}{.3\textwidth}
        \centering
        \includegraphics[width=0.9\linewidth]{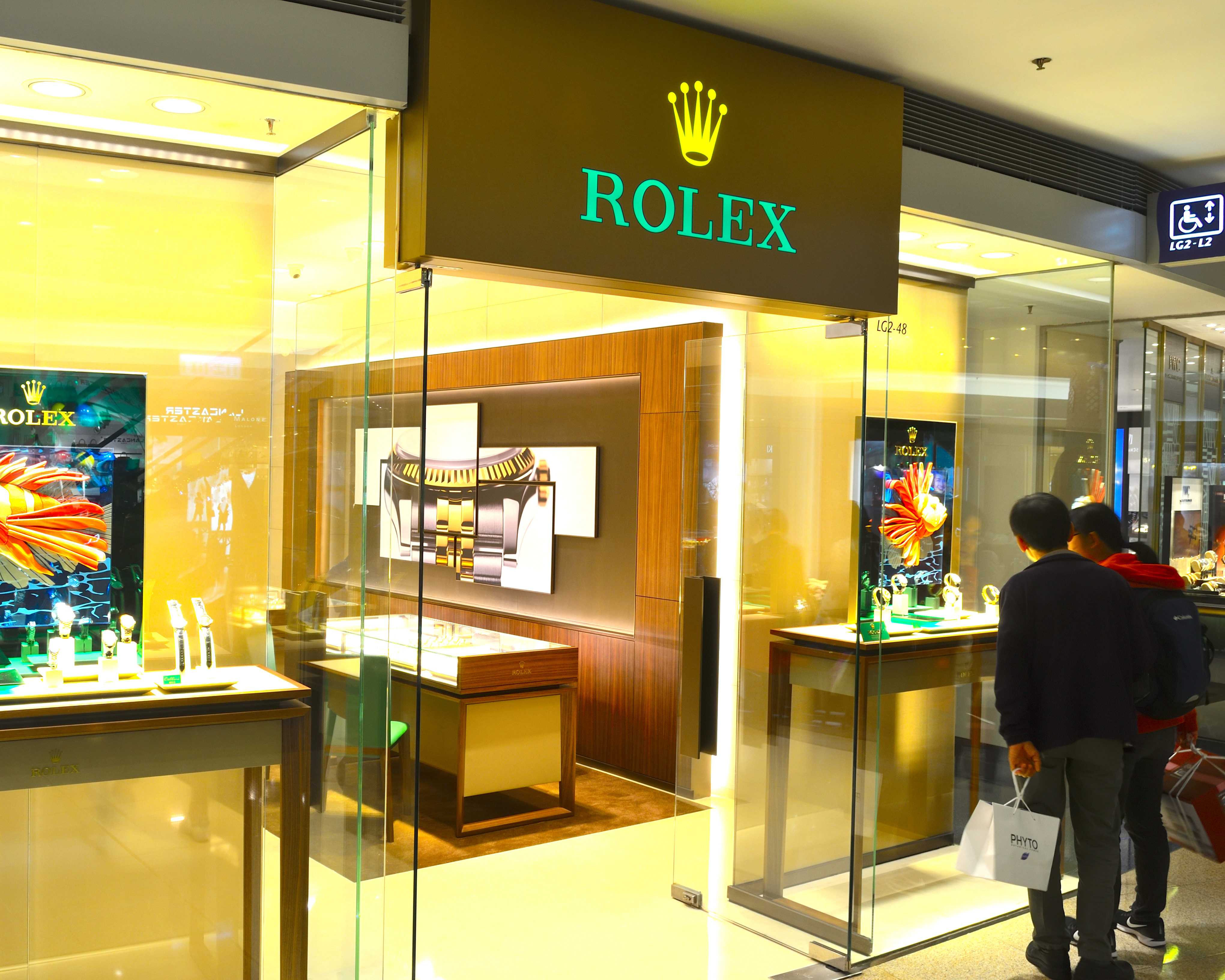} 
        \caption{\tiny{\textit{\textbf{Query}}: Two people showing an interest to purchase a watch.
            \newline
            \textbf{Commonsense}: People looking into the display of a watch store implies they could be interested to purchase a watch there.
            \newline
             \textbf{Visual named entity:} Rolex}
            \newline
            \textbf{Fact}: The store Rolex sells watches.
            \newline
            }
        \label{fig:cofar-first}
    \end{subfigure}
\caption{\label{fig:COFAR-samples}A selection of examples from \data{} showing query, relevant image, associated visual named entity, commonsense and fact.}
\end{figure*}

\section{Related Work}
\label{sec:relwork}
\subsection{Image Search by Visio-lingual alignment} The performance of image search using natural language query has been significantly improved in the last few years. Typically, the methods in this space learn the semantic visio-lingual (V-L) alignment; during retrieval, rank the images according to the learned similarity function. Early works~\cite{faghri2018vse++, Wang_2016_CVPR} learn to project image representations and text embeddings into a joint space. Recently, multimodal transformers have become a de facto model for V-L tasks. Their different avatars~\cite{zhang2021vinvl,lu2019vilbert} tackle multiple V-L tasks jointly by using multi-headed self-attention to encode word tokens and visual objects and are the current state of the art for text-to-image retrieval. However, these methods focus only on the visual cues to represent images and do not encode any external knowledge in their framework. Consequently, any explicit crucial information associated with the image is also ignored.

\subsection{Commonsense and Factual Reasoning}
Bringing commonsense in vision and language tasks is one of the exciting areas of research. The works in this area primarily address: (i) tasks where commonsense reasoning is purely visio-lingual data-driven~\cite{gd-vcr,VisualCOMET,zellers2019vcr,KM-BART} and (ii) tasks where commonsense is enabled by associating the images with external knowledge~\cite{FVQA_wang_2017,OKVQA2019,Marino_2021_CVPR,KVQA2019,textKVQA,AMA_Wu_2016}. Our proposed task falls in the latter category. However, it is distinctly different from others as none of these works address \emph{image search} requiring detailed visual, commonsense as well as factual reasoning \emph{associated to a diverse set of named entities appearing in the image} including business brands, celebrities, and landmarks. Concerning using named visual entities and associated factual reasoning, the only works closest to ours are~\cite{KVQA2019,textKVQA}. However, compared to ours, these works restrict themselves to only celebrities or business brands and have weaker annotations for visual and commonsense reasoning. Despite its importance and many real-world applications on the Web such as news-search, named visual entity linking and its utility towards downstream tasks have been under-explored in the literature. We aim to fill this gap.

\begin{table}[t]
\resizebox{\columnwidth}{!}{
\centering
\begin{tabular}{lr}
    \textbf{\data{} in brief:}&\\
    \toprule
    Number of queries & 40,757 \\
    Number of images & 25,297 \\
    Number of unique named entities & 5,060 \\
    Source of images & text-KVQA~\cite{textKVQA},\\
    & Celebrity in Places~\cite{Zhong16CelebritiesInPlaces},\\
    & Google Landmarks~\cite{weyand2020GoogleLandmarks}.\\
    External knowledge source & Wikipedia\\
    Average query length (words) & 10.5 \\
    Average knowledge length (words) & 43.7 \\
    
\bottomrule\end{tabular}}
\caption{\label{tab:dataset_statistics}\data{} dataset statistics.}
\end{table}

\section{\data{}: Dataset for Image Search requiring \underline{CO}mmonsense and \underline{FA}ctual \underline{R}easoning}
\label{sec:COFAR}
We introduce \data{}, a dataset for studying the novel problem of image search that requires commonsense and factual reasoning. A detailed comparison with related datasets is made in Table~\ref{tab:datasetComp}. \data{} contains images of natural scenes that include visual named entities of business brands, celebrities, and world landmarks. We provide annotations created to query commonsense and factual knowledge pertaining to named entities present in images. We use Wikipedia articles as the external knowledge source for the visual named entities. The dataset contains 40,757 manually annotated English language search queries for 25,297 natural images covering a diverse set of 5,060 named entities. We further provide external knowledge sources for each visual entity. \data{} is made publicly available for download: \url{https://vl2g.github.io/projects/cofar}.

{
\renewcommand{\arraystretch}{1}
\begin{table*}[h!]
    \centering
    \resizebox{1\textwidth}{!}{
      \begin{tabular}{l c c c c c c}
        \toprule
        \textbf{Dataset} &  \textbf{\#Images} &  \textbf{Visual Reasoning} & \textbf{Commonsense Reasoning} & \textbf{Factual Reasoning} & \textbf{Contains Named Entities} & \textbf{External Knowledge} \\
        \midrule
        \textbf{VQA datasets}& &  & & & &\\
        ~~~~FVQA~\cite{FVQA_wang_2017} & 2.1K &  Minimal & Not a major focus & Yes* & \xmark  & Conceptnet\\
        ~~~~KVQA~\cite{KVQA2019} & 24K & Minimal & Not a major focus & Yes & \cmark  & Wikidata\\
        ~~~~text-KVQA~\cite{textKVQA}  & 257K & Minimal & Not a major focus & Yes & \cmark & Wikidata\\
        ~~~~OK-VQA~\cite{OKVQA2019} & 14K & Minimal & Not a major focus & Yes* & \xmark  & Wikipedia \\
       ~~~~VCR~\cite{zellers2019vcr} & 110k & Detailed & Major Focus & No & \xmark & \xmark \\
       ~~~~GD-VCR~\cite{gd-vcr} & 328 & Detailed & \makecell[c]{Major Focus \\ (geo-diverse)} & No & \xmark & \xmark \\
        \midrule
        \textbf{Image search datasets}& &  & & &  &\\
        ~~~~MS-COCO~\cite{lin2014microsoftCOCO} & 120K &
         Detailed & Not a major focus & No & \xmark  & \xmark\\
        ~~~~Flickr30k~\cite{flickr_30_21014} & 30K  & Detailed & Not a major focus & No  & \xmark  & \xmark\\
        ~~~~\textbf{COFAR (This work)} & \textbf{25K} & \textbf{Detailed} & \textbf{Major focus} & \textbf{Major Focus} & \textbf{\cmark}  & \textbf{Wikipedia}\\
 
       \bottomrule
      \end{tabular}}
      \caption{\label{tab:datasetComp}Comparison of \data{} with other related datasets. Examples of Minimal vs. Detailed visual reasoning: `How many chromosomes does the creature in this image have?' (Source: OK-VQA) vs. `\textbf{A lady wearing a blue t-shirt} going home after purchasing groceries' (Source: \data{}). Further, Yes* under the factual reasoning column indicates that though these datasets require factual reasoning, their facts are about common objects (such as Orange is a citric fruit) and not about named entities (such as Lionel Messi is an Argentine professional footballer). Besides detailed visual reasoning, commonsense and factual reasoning associated with \emph{visual named entities} appearing in the image are unique aspects of \data{} that distinguish it from other related datasets.}
\end{table*}
}

\subsection{Image collection:} 
\begin{figure}[t]
    \centering    \includegraphics[width=0.9\columnwidth]{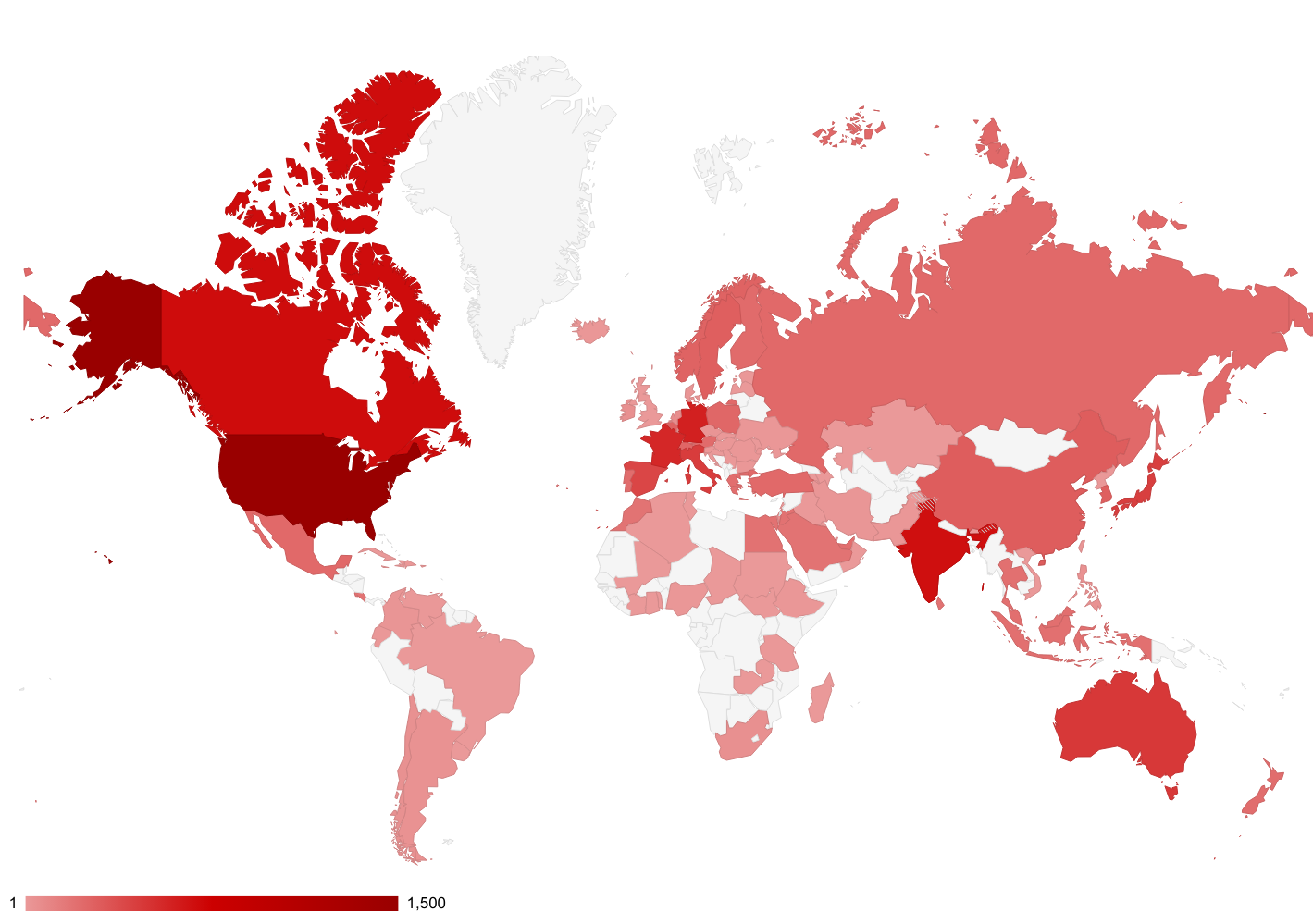}
     \caption{\label{fig:cofar-geo}Distribution of named entities in \data{} on the world map. \data{} contains named entities from a diverse list of countries, with a slight unintentional bias towards countries such as the United States of America and Canada. Darker color indicates more entities.}
\end{figure}

\begin{figure}[!t]
    \centering
    \includegraphics[width=1.0\columnwidth]{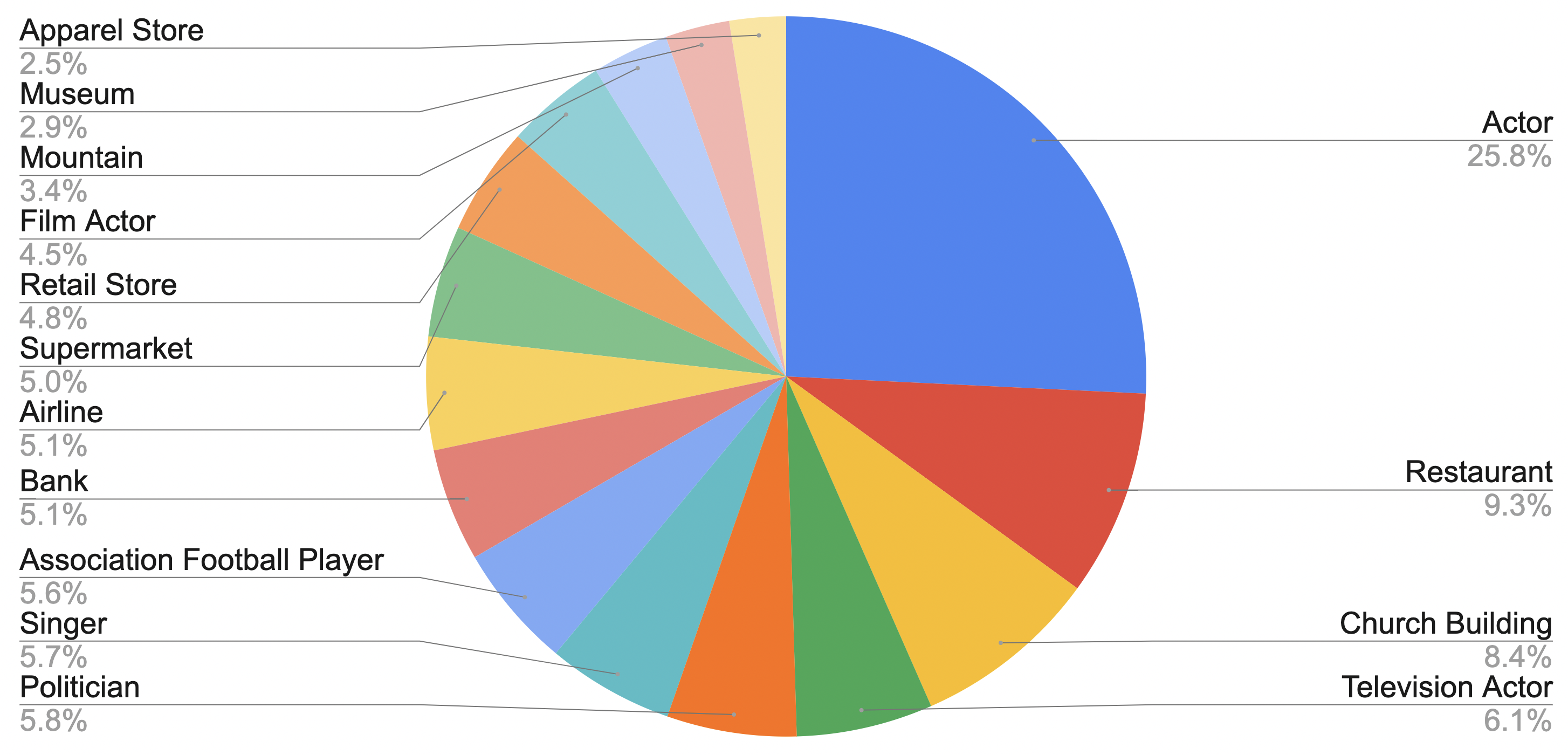}
    \caption{\label{fig:cofar-type}Distribution of the top fifteen categories of named entities present in \data{}.}
\end{figure}

We begin our dataset creation process by collecting images containing one of the three popular named visual entity types: business brands, famous personalities, and landmarks across the globe. To this end, we first started collecting images from different publicly available sources, i.e., we obtain natural scene images containing business brands, personalities, and landmarks using text-KVQA~\cite{textKVQA}, VGG-celebrity in places~\cite{Zhong16CelebritiesInPlaces} and the Google landmarks~\cite{weyand2020GoogleLandmarks} respectively.\footnote{Restricted by the budget, instead of choosing entire celebrity in places and the Google landmarks, we choose a reasonably large subset.} Note that these sources do not provide any natural language queries relevant to the images and, therefore are not directly usable for our task. 
We then associate each of these images with the Wikipedia page of the entity it contains. Note that during training, this association is assumed to be known, but during testing, we perform visual entity linking. Some of the example entities in our dataset are \emph{Rolex}, \emph{Lionel Messi}, and the \emph{Eiffel Tower}. As shown in Figure~\ref{fig:cofar-geo} the distribution of visual named entities in the images of our dataset is geographically diverse. Further, we also illustrate the diversity in the category-wise distribution of \data{} in Figure~\ref{fig:cofar-type}. We refer the reader to the Appendix for further details on \data{}.

\subsection{Manual annotation:} The images, along with their associated Wikipedia summary texts, were given to three hired human annotators with the task of annotating queries. These annotators were from geographically diverse locations and had proficiency in written English. In particular, they were instructed to create queries that include (i) factual information of the entity present in the image, for example, \emph{captain of the Argentina national football team, landmark located in Paris}, as well as (ii) commonsense knowledge about events, activities, people, what is going to happen in the scene, or what might have just occurred, for example, \emph{celebrating after scoring a goal, people in the image are getting married}. Annotators have also been given the option to discard those images where it is very hard to associate visual commonsense, for example, just a frontal view image of a landmark or a signboard of a business brand or an image without any interesting visual activity around. The entire process of manually coming up with queries that require commonsense and factual reasoning, followed by a manual quality check of the data, took approximately 800 person-hours by three annotators. At the end of this stage, we obtained 25K images and 40K queries involving commonsense and factual information about the image. Table~\ref{tab:dataset_statistics} summarizes the dataset statistics of \data{}.

A selection of examples from \data{} are shown in Figure~\ref{fig:COFAR-samples}. An image search model relying exclusively on visual cues would find it challenging to retrieve the relevant images for the queries in~\data{}. Consider search query-(c) shown in the figure i.e., two people showing interest in purchasing a watch.. In this image, two people are looking at a display in a Rolex store that sells watches (world knowledge). Therefore, even though detecting watches in this image may be hard for vision models, the matching image shown at the top of this query is relevant. The use of visual entity recognition to associate encyclopedic knowledge and commonsense and factual reasoning are some of the salient features that make \data{} distinctly different from existing text-to-image retrieval datasets. 

\subsection{Train and Gallery Split:}
Based on categories of named entities present, dataset is grouped into \data{} (landmark), \data{} (celeb), and \data{} (brand). All the baselines and our proposed method are evaluated on them separately as well together. Further, we split the dataset into (i) \textbf{Train set:} Used for learning image-query alignment, this set contains 12,120 images and 33,800 queries. (ii) \textbf{Small and large gallery sets:} We show retrieval on two gallery sets containing 1K and 5K images for \data{}. We use 2,800, and 9,800 natural language queries in all for 1K and 5K image galleries, respectively. Please note that retrieval on the test galleries is performed with images containing \emph{entities that are unseen} during training.

\section{\underline{K}nowledge \underline{R}etrieval-\underline{A}ugmented \underline{M}ultimodal \underline{T}ransformer (\model{})}
\label{sec:approach}
\begin{figure*}[!t]
\centering
\includegraphics[width=0.9\textwidth]{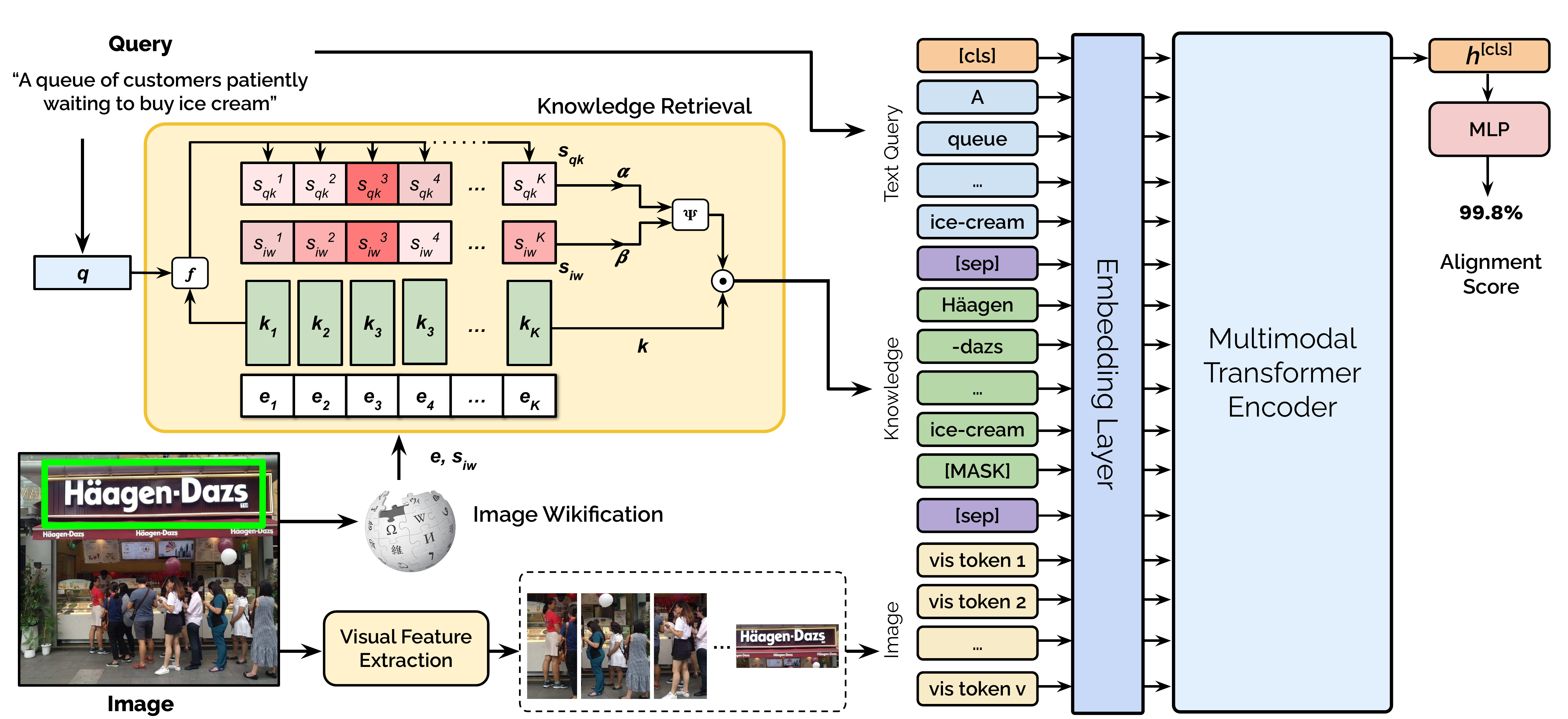}
\caption{\label{fig:kmmt}\textbf{Overview of proposed Knowledge Retrieval Augmented Multimodal Transformer (\model{}):} Given a query and a ranked list of visual entities identified in the image, \model{} grounds the relevant knowledge. This grounded knowledge, along with visual objects and natural query, is fed to a multimodal transformer that learns to align query and relevant image. Please refer Section~\ref{sec:approach} for more details. \textbf{[Best viewed in color]}.}
\end{figure*}
Given a natural language query and a large gallery of images each containing a visual named entity, our goal is to retrieve relevant images. To this end, we present \underline{K}nowledge \underline{R}etrieval-\underline{A}ugmented \underline{M}ultimodal \underline{T}ransformer (\model{}) -- an unified framework that contains two major modules: (i) visual entity and query-aware knowledge retrieval and (ii) knowledge-infused multimodal transformer as illustrated in Figure~\ref{fig:kmmt}.

\subsection{Visual Entity and Query-Aware Knowledge Retrieval:}
We posit that visual entities appearing in the image act as a gateway to the encyclopedic knowledge, and its integration to an image retrieval system has the potential to bring commonsense and factual reasoning ability. Therefore,  
to associate visual entities appearing in the given image to their corresponding Wikipedia page, we perform \emph{visual entity linking} or \iw{} which is an analogous task to Wikification~\cite{fast-e2e-wikification} of text corpora, i.e. linking entity mentions in text documents to their corresponding Wikipedia page. More formally, given an image, a set of $m$ candidate entities ${\cal E}=\{e_1,e_2,\cdots,e_m\}$ containing business brands, celebrities, and world landmarks, and associated knowledge text (obtained from Wikipedia articles of these entities) ${\cal K}=\{k_1, k_2, \cdots, k_m\}$; \iw{} aims to rank these entities with respect to their image wikification likelihood ($s_{iw}$). Here, for an image, $s^u_{iw}$ denotes likelihood of $u$th entity in that image. 
We obtain these likelihood scores by using off-the-shelf approaches such as CRAFT+CRNN~\cite{CRAFT_2019,CRNN_2017} for detecting and recognizing business brand mentions in the image, VGG face~\cite{VGGFace} for comparing celebrity faces appearing in the images against a set of reference faces, and landmark recognition~\cite{weyand2020GoogleLandmarks} for recognizing world landmarks. 

If we link images to only that entity which corresponds to the highest likelihood score, linking may be incorrect (especially due to look-alike faces or similar world landmarks or noisy text recognition). This is also evident from the experiment, which clearly shows the gap between top-1 and top-K performance of visual entity linking (Refer to Table~\ref{tab:wikification_res}). To resolve any error in visual entity linking and subsequently retrieving relevant knowledge, we further leverage the natural language query. To this end, we compute the similarity between query and knowledge text associated with top-K entities using a trainable BERT model $f$ and denote these similarity scores as $s_{qk}$ where $s^u_{qk}$ denotes the similarity between query and knowledge text corresponding to $u${th} entity. Further, relevance of each entity with respect to image and given query is computed as follows: $s=\Psi(\alpha s_{iw} + \beta s_{qk})$, here $\Psi$ is argmax. The choice of argmax over softmax is intuitive as only one knowledge text is relevant for a given query and image in our task. Once we obtain $s$, we perform element-wise multiplication to ${\cal K}=\{k_1, k_2 \cdots k_K\}$ and feed this knowledge to a multimodal transfer as described next.

\subsection{Knowledge-infused Multimodal Transformer:}
Once we obtain relevant knowledge from our knowledge retrieval module,  
we use Knowledge-infused Multimodal Transformer -- a simple and effective architecture to learn alignment between natural language search queries and images along with their associated external knowledge. \model{} seamlessly integrates these three input modalities in a unified end-to-end trainable architecture. To achieve this, we first encode the query text, knowledge text, and visual regions as three sequences of features. We then project these features to a shared embedding space before using them as input to the \model{}. These features then attend to each other through multiple self-attention layers ~\cite{vaswani2017attention}. The output of a special class token from the final layer's output is then used to predict the alignment between the query and image along with its knowledge text.   

\subsection{Pretraining:}
We learn a strong vision-language grounding capability in \model{} through pretraining on MS-COCO~\cite{lin2014microsoftCOCO} with the objective tasks of masked language modelling (MLM) and image text matching (ITM).

\subsection{Query and Knowledge Encoder:} We fine-tune pretrained BERT~\cite{BERT_2019} to encode the text of the query and external knowledge. For a given search query $Q$ containing $L$ words and a given knowledge $k_i$ containing $M$ words, we embed them into sequences of $d$-dimensional BERT feature vectors $\{q_l\}_{l=1}^{L}$ and $\{k_{ij}\}_{j=1}^{M}$ respectively.

\subsection{Image Encoder:} Given an image, we detect a fixed set of $N$ visual objects using Faster R-CNN \cite{FasterRCNN2015} pretrained on Visual Genome~\cite{krishnavisualgenome}. Each image $I$ is represented as an unordered sequence of the $N$ object proposals $\{R_i\}_{i=1}^{N}$ where each $R_i$ is represented as (${R_i^{cnn}}$, ${R_i^{bbox}}$), which denote \emph{2048}-dimensional region feature and \emph{4}-dimensional spatial feature, respectively.

We project regional feature ${R_i^{cnn}}$ and spatial feature ${R_i^{bbox}}$ into the same $d$-dimensional space as the search query and the knowledge text using two different learnable transformation matrices $\mathbf{W}_{cnn}$ and $\mathbf{W}_{bbox}$. 
We apply layer normalization $L(\cdot)$~\cite{LayerNorm_2016} to each transformed feature, and add them to get the final visual object feature ${F_{R_i}}$.
\begin{equation}
{F_{R_i}} = L(\mathbf{W}_{cnn}R_{i}^{cnn}) + L(\mathbf{W}_{bbox}R_{i}^{bbox}).
\end{equation}

\subsection{Query-Image Alignment Learning:}
Besides learning $d$-dimensional embeddings for the three inputs, we also learn it for three special tokens, namely {[$SEP$]} to separate the input modalities, {[$CLS$]} to calculate the final alignment score and {[$MASK$]} to replace the text tokens during MLM. We then allow all the $L + M + N + 3$ input token features to attend to each other through $T$ transformer encoder layers to obtain a joint representation. 

As the final step, a multi-layer perceptron that takes $d$-dimensional $[CLS]$ output feature and produces an alignment score $Out^{[CLS]}$ indicating if the given pair of a search query and the image with associated knowledge are aligned or not, is used. During training, we create positive pairs by selecting images and their corresponding queries from the dataset and negative pairs by randomly changing either the image or the query of the selected pair with another random choice in the dataset. We train the model using binary classification loss. Further, to make the image-query alignment robust, we also train the model with the MLM objective wherein each iteration of training, we replace text input tokens at random with a special token {[$MASK$]} with a probability of 0.15 and predict the masked tokens based on the context of image, query, and knowledge.
During retrieval, for a given query, we rank all the images in the gallery based on the predicted alignment scores. Further implementation details of \model{} are provided in the Appendix.
\begin{table*}[t]
  \centering
  \resizebox{1\textwidth}{!}{
  \begin{tabular}{l c c c c c c c c c c c c c c c c c}
  \toprule
  \multicolumn{1}{c}{} & \multicolumn{4}{c}{\data{} (Unified)} & \multicolumn{4}{c}{\data{} (Brand)} & \multicolumn{4}{c}{\data{} (Celeb)} &
  \multicolumn{4}{c}{\data{} (Landmark)}  \\
  
  \cmidrule(r){2-5}
  \cmidrule(r){6-9}
  \cmidrule(r){10-13}
  \cmidrule(r){14-18}
   
  Method & R1 & R5 & R10 & MdR & R1 & R5 & R10 & MdR & R1 & R5 & R10 & MdR & R1 & R5 & R10 & MdR\\
  \midrule
  \multicolumn{18}{c}{\cellcolor[gray]{0.8}{\textbf{1K Gallery}}}\\
  \midrule
  \textbf{Knowledge-only} & & & & & & & & & & & & \\
   ~~~~Sentence similarity & 3.1 & 8.7 & 19.0 & 84 & 2.4 & 9.3 & 18.8 & 68 & 3.0 & 8.2 & 16.9 & 143 & 4.2 & 9.1 & 19.3 & 97 \\
  \midrule
   \textbf{Vision-only} & & & & & & & & & & & & & & & & & \\
   ~~~~VSE++~\cite{faghri2018vse++} & 7.4 & 19.2 & 23.8 & 68 & 6.9 & 19.5 & 27.6 & 60 & 6.0 & 25.1 & 38.5 & 27 & 21.8 & 48.0 & 59.0 & 9 \\
   ~~~~VisualBERT~\cite{li-etal-2020-bert-vision} & 22.7 & 50.0 & 62.5 & 5 & 24.0 & 50.9 & 63.3 & 5 & 8.0 & 29.3 & 37.3 & 22 & 32.4 & 64.5 & 70.0 & 4 \\
   ~~~~ViLBERT~\cite{lu2019vilbert} & 29.8 & 57.9 & 71.0 & 5 & 28.1 & 55.4 & 68.6 & 4 & 16.5 & 34.4 & 42.0 & 15 & 36.0 & 66.9 & 74.0 & 4 \\
   ~~~~VinVL~\cite{zhang2021vinvl} & 30.5 & 62.1 & 74.3 & 4 & 31.2 & 64.8 & 75.7 & 4 & 18.3 & 38.9 & 46.5 & 10 & 38.7 & 68.0 & 76.3 & 3 \\
   \midrule
   \textbf{Knowledge-aware V-L Models} & & & & & & & & & & & & & & & & \\
    ~~~~Modified Memory Network & 15.2 & 35.0 & 50.3 & 5 & 14.4 & 34.9 & 48.6 & 18 & 6.1 & 26.8 & 39.4 & 23 & 24.5 & 51.1 & 60.3 & 5 \\
    ~~~~KQIA & 22.0 & 52.4 & 64.5 & 5 & 19.9 & 48.2 & 57.5 & 9 & 10.1 & 29.2 & 40.5 & 19 & 31.9 & 57.8 & 67.0 & 5 \\
    ~~~~KRISP-inspired model & 28.1 & 53.8 & 69.0 & 4 & 26.8 & 51.5 & 67.6 & 5 & 13.6 & 32.5 & 39.8 & 17 & 34.3 & 65.9 & 74.2 & 3 \\
    ~~~~\textbf{Ours} &  &  &  &  &  &  &  &  &  &  & &  &  &  &  & \\
     ~~~~~~~~\textbf{\model{} (w/o Vision)} & 1.9 & 6.6 & 12.6 & 57 & 1.1 & 7.4 & 12.4 & 35 & 2.6 & 6.6 & 17.1 & 164 & 2.7 & 10.9 & 14.5 & 100 \\
    ~~~~~~~~\textbf{\model{} (w/o Knowledge)} & 19.8 & 39.1 & 49.8 & 14 & 19.4 & 38.3 & 49 & 15 & 11.8 & 26.3 & 35.5 & 25 & 35.5 & 67.3 & 74.5 & 2 \\
    ~~~~~~~~\textbf{\model{}} & \textbf{31.6} & \textbf{64.4} & \textbf{76.2} & \textbf{3} & \textbf{32.9} & \textbf{66.5} & \textbf{78.6} & \textbf{3} & \textbf{19.7} & \textbf{44.7} & \textbf{51.3} & \textbf{8} & \textbf{40.0} & \textbf{69.1} & \textbf{80.0} & \textbf{2} \\
    ~~~~~~~~\textbf{\model{} (Oracle)} & 40.0 & 73.2 & 84.5 & 2 & 38.5 & 72.0 & 83.3 & 2 & 26.3 & 48.7 & 61.8 & 6 & 42.7 & 76.4 & 87.3 & 2\\
    \midrule
    \multicolumn{18}{c}{\cellcolor[gray]{0.8}{\textbf{5K Gallery}}}\\
    \midrule
     \textbf{Vision-only} & & & & & & & & & & & & & & & & \\
   ~~~~VSE++~\cite{faghri2018vse++} & 4.7 & 11.2 & 18.0 & 119 & 3.9 & 9.2 & 17.4 & 128 & 2.9 & 9.1 & 12.5 & 274 & 8.8 & 20.4 & 33.6 & 49 \\

   ~~~~VisualBERT~\cite{li-etal-2020-bert-vision} & 11.4 & 28.6 & 40.0 & 19 & 11.1 & 28.0 & 38.8 & 20 & 6.7 & 13.3 & 20.0 & 95 & 13.6 & 31.0 & 40.1 & 18 \\
   ~~~~ViLBERT~\cite{lu2019vilbert} & 13.6 & 31.7 & 43.5 & 12 & 13.0 & 30.8 & 41.5 & 10 & 9.1 & 15.8 & 25.0 & 67 & 12.2 & 43.6 & 54.0 & 8 \\
   ~~~~VinVL~\cite{zhang2021vinvl} & 15.9 & 35.6 & 49.2 & 10 & 14.9 & 33.6 & 44.5 & 9 & 11.2 & 17.7 & 30.4 & 31 & 14.2 & 44.9 & 58.0 & 6 \\
   \midrule
   \textbf{Knowledge-aware V-L Models} & & & & & & & & & & & & & & & & \\
    ~~~~Modified Memory Network & 7.3 & 21.8 & 34.6 & 40 & 6.8 & 19.9 & 30.1 & 46 & 3.8 & 10.1 & 14.6 & 143 & 9.3 & 26.8 & 37.9 & 38\\
    ~~~~KQIA & 9.8 & 25.3 & 36.2 & 21 & 9.1 & 24.9 & 35.4 & 24 & 7.7 & 14.9 & 20.8 & 79 & 10.8 & 28.1 & 37.4 & 28 \\
    ~~~~KRISP-inspired model & 14.1 & 36.6 & 45.9 & 10 & 13.3 & 32.4 & 43.7 & 10 & 8.8 & 14.1 & 23.9 & 61 & 12.0 & 41.4 & 53.7 & 7 \\
    ~~~~\textbf{Ours} &  &  &  &  &  &  &  &  &  &  & &  &  &  &  & \\
    ~~~~~~~~\textbf{\model{}} & \textbf{17.1} & \textbf{42.9} & \textbf{57.2} & \textbf{8} & \textbf{16.7} & \textbf{42.2} & \textbf{56.5} & \textbf{8} & \textbf{11.8} & \textbf{18.4} & \textbf{34.2} & \textbf{28} & \textbf{12.7} & \textbf{45.5} & \textbf{58.2} & \textbf{6} \\
    ~~~~~~~~\textbf{\model{} (Oracle)} & 18.9 & 45.8 & 59.9 & 8 & 18.5 & 45.0 & 58.9 & 7 & 15.8 & 25 & 38.2 & 18 & 18.2 & 52.7 & 65.5 & 5  \\
  \bottomrule
\end{tabular}}
\caption{\label{tab:mainRes-1}Comparison of retrieval performance on \data{} (with 1K and 5K gallery each) with baselines and ablations. We report mean recall (R) at top 1, 5, and, 10 retrievals and median rank (MdR) over all the test queries.}
\end{table*}

\section{Experiments and Results}
\label{sec:expts}
\label{sec:exptBaselines}
We group image retrieval baseline approaches into three categories: (i)  Knowledge-only, (ii) Vision-only, and (iii) Knowledge-aware vision and language (V-L) models to investigate the following questions respectively:
\begin{itemize}[noitemsep]
    \item How much impact does external knowledge have? Can it alone drive performance in \data{} without any visual cues?
    \item Is there a need for integrating external knowledge in \data{}?
    \item How do other knowledge-aware baselines perform on \data{}?
\end{itemize}

Under \textbf{Knowledge-only}, we utilize BERT~\cite{BERT_2019} to perform query-knowledge sentence-matching. In \textbf{VL models}, we use modern text-to-image retrieval methods, namely VSE++~\cite{faghri2018vse++}, and competitive vision-and-language transformers such as VisualBERT~\cite{li-etal-2020-bert-vision}, ViLBERT~\cite{lu2019vilbert}, and VinVL~\cite{zhang2021vinvl}. 
\begin{figure*}[!t]
\centering
\includegraphics[width=1\textwidth]{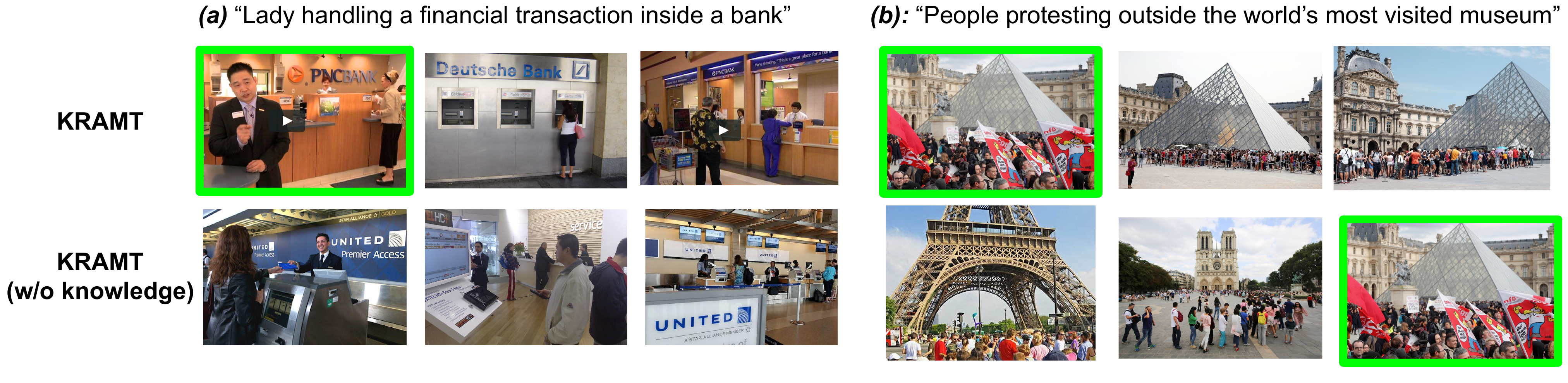}
\caption{\label{fig:visRes} Top-3 retrieved images using proposed \model{}(w/o Knowledge) and \model{} on \data{}-1K for two queries. We see that models without access to external knowledge often fail to interpret commonsense such as a financial transaction or protest, and factual information, such as the world's most visited museum, present in the query. On the contrary, \model{} retrieves semantically more coherent images. Here green colored bounding box indicates the ground truth image.}
\end{figure*}
\textbf{Knowledge-aware VL models:}
As there are no directly comparable knowledge-aware image-retrieval methods in current literature, we implement a few knowledge-aware visual question answering-based models with appropriate modifications to make them compatible for our task: \textbf{(i) Modified Memory Network:} Memory networks, and their variations have shown to yield state-of-the-art performance on knowledge-aware VQA benchmarks~\cite{KVQA2019,vkmn}. We implement this baseline by using top-K knowledge texts. These texts are scored with a query, and the weighted sum of this representation, CNN features of the image, and query representation are passed to a binary classifier that classifies if the image is relevant to the query.
\textbf{(ii) KRISP-inspired model:} KRISP~\cite{Marino_2021_CVPR} addresses open knowledge-based VQA using implicit and symbolic knowledge stored in a graph data structure. In our setting, we use unstructured knowledge text in place of symbolic knowledge. We model implicit knowledge using MM-BERT, similar to KRISP, and for unstructured text, we use BERT embedding of the knowledge text. The output of these representations along with BERT-based query representation is fed to an MLP for learning alignment. 
\textbf{(iii) KQIA}: Here, knowledge text, along with queries and images, are encoded using gated recurrent units and CNN, respectively, and are then projected into a common space to learn alignment. All baselines are pretrained on the COCO dataset unless mentioned otherwise.

\subsection{Ablations:}
To evaluate the effect of different components of \model{}, we present the following ablations: \textbf{\model{} (w/o Knowledge):} where knowledge text is omitted, 
\textbf{\model{} (w/o vision):}  where only query and retrieved knowledge is used, and \textbf{\model{} (Oracle)} that assumes  ground-truth knowledge is available to the model. 

\begin{table}[t]
    \centering
    \resizebox{1\columnwidth}{!}{
      \begin{tabular}{l c r r r r}
        \toprule
        \multicolumn{1}{c}{} & \multicolumn{1}{c}{\# of} & \multicolumn{4}{c}{\data{}-1K} \\
        \cmidrule(r){3-6}
        Method & Pre-train Images & R1 & R5 & R10 & MdR \\
        \midrule
        CLIP~\cite{radford2021learning} & 400M & 26.4 & 58.1 & 72.8 & 6 \\
        12-in-1~\cite{Lu_2020_CVPR} & 6.3M & 30.2 & 59.9 & 74.3 & 4 \\
        \model{} & 125K & \textbf{31.6} & \textbf{64.4} & \textbf{76.2} & \textbf{3}\\
        \bottomrule
      \end{tabular}}
      \caption{\label{tab:large_pretrain_Res}Using external knowledge over very large-scale pretraining on \data{}~1K.}
\end{table}
\subsection{Results and Discussions}
\label{sec:exptRes}
We quantitatively evaluate \model{} on \data{} and compare it against related approaches in Table~\ref{tab:mainRes-1}. We report recall (R1, R5 and, R10) and median rank (MdR) averaged over all the test queries. Note that higher values for recall and lower values for median rank are desired. The poor performance of knowledge-only models confirms that image search in \data{} is non-trivial and external knowledge about the entities in images alone is insufficient. Further, we observe that the vision-only models such as VisualBERT, ViLBERT, and VinVL, without access to external knowledge, do reasonably well solely through visual reasoning. However, it falls short to \model{}. By virtue of its seamless integration of search query, visual content, and unstructured knowledge, \model{} clearly outperforms other baselines, including other Knowledge-aware V-L baselines. These results show the effectiveness of transformer-based methods in \data{} task. 
The results of ablations are also reported in Table~\ref{tab:mainRes-1}. Here, we observe that \model{} that leverages harvested knowledge for enabling commonsense and factual reasoning is significantly superior to \model{} (w/o knowledge).

\subsection{Models Pretrained on large-scale datasets} We note it may not be fair to compare our model with those which use very-large-scale datasets for pretraining due to significant differences in size of training data. Moreover, there is possibility of overlap of images in their train sets and \data{}-test set; for the sake of a comprehensive comparison, we compare \model{} with two modern transformer-based models namely CLIP~\cite{radford2021learning} and 12-in-1~\cite{Lu_2020_CVPR} in Table~\ref{tab:large_pretrain_Res}. Please note that they use 400M and 6.3M images, respectively, for pretraining as compared to 125K images (COCO) in our model. We see \model{} surpasses CLIP and 12-in-1 despite being a smaller model.

We show a selection of visual results for top-3 retrievals for two queries in Figure~\ref{fig:visRes}. The retrieved images by \model{} (w/o knowledge) may contain the relevant image, but often ranked lower due to their inability to recognize the entities and perform factual reasoning. On the contrary, the proposed \model{} consistently retrieves relevant images, confirming our hypothesis.

\begin{table}[!t]
    \centering
    \resizebox{0.6\columnwidth}{!}{
      \begin{tabular}{l c r r}
        \toprule
        \data{} Category  & Top 1 (\%) & Top 5 (\%) \\
        \midrule
        Brand & 60.8 & 79.6 \\
        Landmark & 63.5 & 70.2 \\
        Celeb & 80.1 & 83.0 \\
        \bottomrule
      \end{tabular}}
      \caption{\label{tab:wikification_res}Results of \iw{} (visual entity linking) on different categories of \data{} test data.}
\end{table}

\subsection{Limitations and Future Scope} We observe the following limitations of our work: (i) for the introduction of \data{}, we have chosen natural scenes that contain only one visual named entity. This may not be the case in a real-world setting, (ii) restricted by the budget, current version of \data{} contains only 25K images of 5K named entities in all. However, in an open-set scenario, a much larger and diverse set of visual named entities can be considered, and \iw{} can be a promising research challenge. In fact a contemporary work~\cite{visualEntityLinking} poses this as a stand-alone task, and (iii) explicit external knowledge associated with common objects has not been leveraged. We leave addressing these limitations as a future work of this paper.

\section{Conclusion}
\label{sec:con}
In Information Retrieval and NLP community, knowledge bases are instrumental in enabling commonsense and semantic search. However, their utility in semantic image search has not been extensively explored in the literature. We have drawn the attention of the vision and language community towards this issue through our work and presented a novel multimodal transformer namely \model{} which seamlessly combines image, query, and knowledge encoding to learn alignment between the image with associated knowledge and query. We firmly believe that image search requiring commonsense and factual reasoning and the new dataset viz. \data{} introduced in this work will open up several future research avenues.

\section{Ethical Considerations}
\label{sec:ethics}
One caveat of \data{} is that the images have been collected from various publicly available sources that may contain geographical bias inherently present in them that were undetected in this work. This problem is common with many public vision benchmarks. A more rigorous inspection is indeed required before deploying the proposed model for real-world applications.

\section*{Acknowledgements}
We are grateful to the anonymous reviewers and area chairs for their insightful suggestions and feedback. We thank \href{https://www.accenture.com/in-en/about/accenture-labs-index}{Accenture Labs} for supporting this work.

\bibliography{references}
\bibliographystyle{acl_natbib} 

\clearpage
\appendix
\section*{Appendix}

\subsection*{\model{} Pre-training}
To train our full \model{} model, we initially pretrain on the COCO captions dataset~\cite{lin2014microsoftCOCO} for the objective task of image-caption alignment and masked language modelling. COCO presents a huge diversity of visual content and serves as a good dataset for improving visual reasoning abilities in \model{}. Further, the model is finetuned on the trainset of \data{}.

\subsection*{\model{} Implementation Details}
We implement the code in  PyTorch ~\cite{NEURIPS2019_9015}. The transformer layers of \model{} are implemented using Hugging Face's transformers library~\cite{wolf-etal-2020-transformers}. We use three transformer encoder layers, with 8 attention heads. The hidden dimension of each block of the transformer layer, as well as the input token feature dimension, is the same as the standard BERT~\cite{BERT_2019} model's hidden dimension of 768.

To encode the query, we use pretrained BERT (`bert-base-uncased') provided by Hugging Face. We keep the sequence length of query text to 40, by truncating the longer sequences and padding the shorter ones. To encode knowledge text, we use the same pretrained BERT, however, this time we keep the sequence length to 80 to accommodate the Wikipedia summary of a page (typically at most 70 words long). This BERT is further fine-tuned during the training of \model{} with 0.1 times smaller learning rate than that of the \model{} layers.

To encode images, we extract visual objects using Faster R-CNN~\cite{FasterRCNN2015} pretrained on Visual Genome~\cite{krishnavisualgenome}. We use top-50 most confident visual object proposals for each image, and represent the visual object's appearance features using Faster R-CNN's `fc6' features of 2048 dimensions. For spatial features, we use 4-dimensional normalized bounding box representation as mentioned in our approach in the main paper. To represent special tokens $[CLS]$ and $[SEP]$ we learn 768-dimensional embedding for each of them during training. 

To get alignment scores from the output embedding of the $[CLS]$ token, we learn a multi-layer-perceptron (MLP) with one hidden layer of size $512$ and a ReLU activation. For pretraining on COCO, the knowledge text input is masked and trained for 42 epochs using Adam~\cite{kingma2014adam} optimizer, with a constant learning rate of 1e-4. Before we finetune \model{} on \data{} for the task of query-image alignment, we finetune \model{} on text of \data{} with just masked language modelling objective for 10 epochs using Adam~\cite{kingma2014adam} optimizer, with a constant learning rate of 5e-5. Finally, we finetune \model{} on \data{} with the task of query-image alignment for 15 epochs using Adam~\cite{kingma2014adam} optimizer, with a constant learning rate of 0.00002. The model is trained with the binary cross-entropy loss for query-image alignment task, and cross-entropy loss over vocabulary for masked language modelling task. The model was trained using two Nvidia RTX 5000 GPUs (each having 16GB of GPU memory) with a batch size of 64 while training and 128 while testing. \model{} pretraining takes approximately four days on the two GPUs, whereas \model{} finetuning on \data{} takes lesser time.

Further details of the implementation can be found in the code which we provide in the project page.

\begin{figure}[t]
    \centering
    \includegraphics[width=1.0\columnwidth]{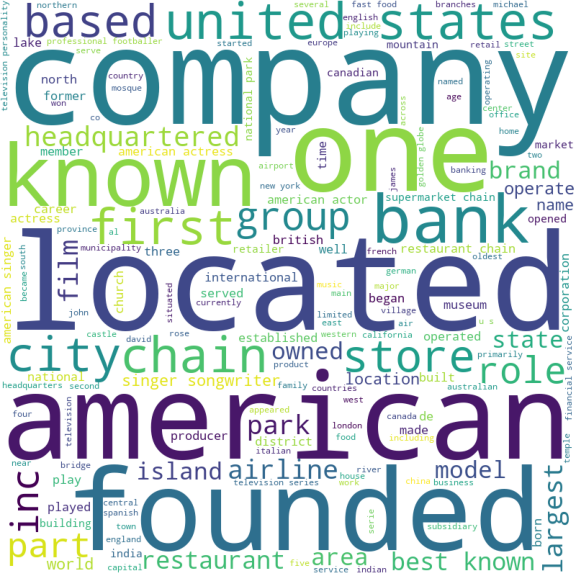}
    \caption{\label{fig:k-wordcloud}Knowledge word cloud}
\end{figure}

\clearpage

\begin{figure*}[!t]
\centering
\includegraphics[width=0.9\textwidth]{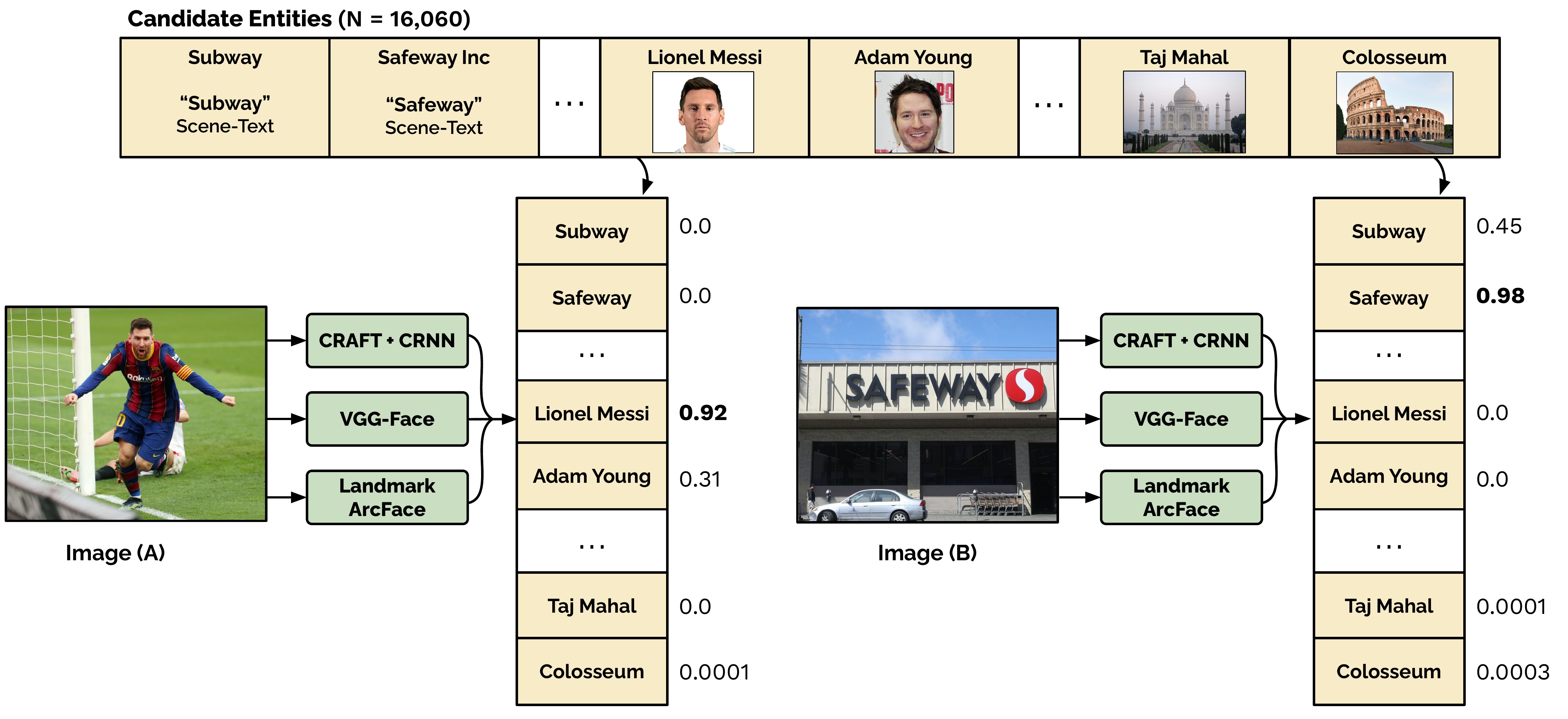}
\caption{\label{fig:image-wikification} \textbf{Overview of \iw{} (visual entity linking) method in \model{}}. To recognize named visual entities in images, we use available methods such as CRAFT+CRNN, VGG-Face, and Landmark ArcFace for brands, celebrities, and landmarks respectively. Using these experts, we measure similarity against several thousands of reference entities to obtain a set of high ranking candidates. This open-set recognition approaches allow for addition or removal of any number of reference entities without a need to re-train.}
\end{figure*}

\begin{figure*}[h]
\centering
\includegraphics[width=0.9\textwidth]{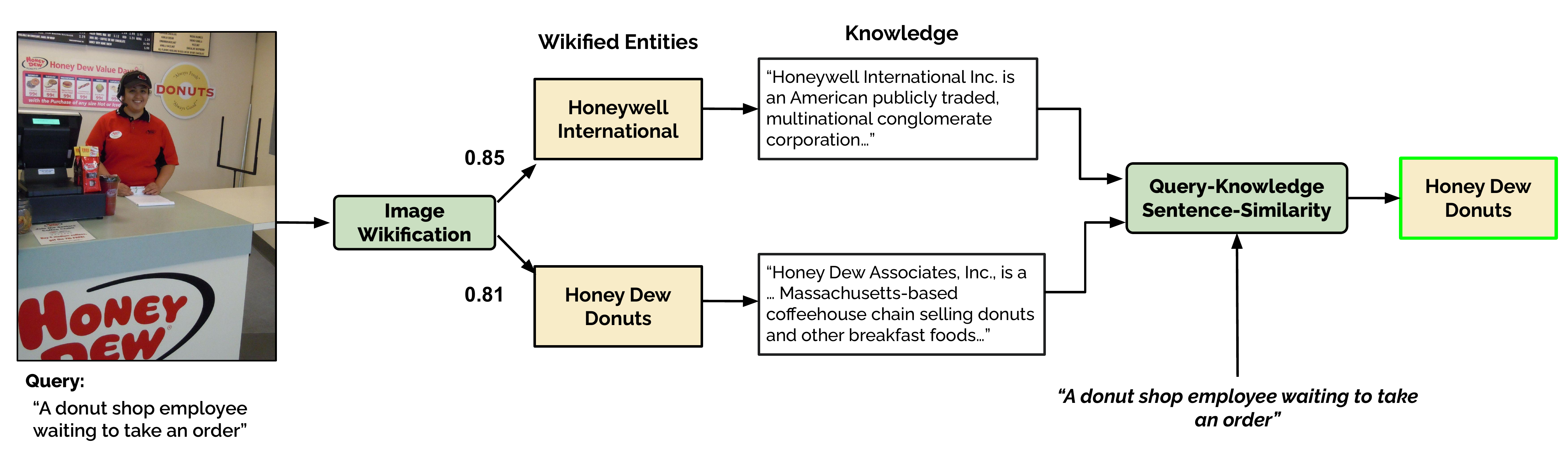}
\caption{\label{fig:query-reranking} \textbf{Using query-based guidance in knowledge-retrieval for \model{}.} Taking the set of top-ranked candidate entities, we use the search query to select the most appropriate entity by measuring sentence-similarity between the query and entity's knowledge text.}
\end{figure*}

\begin{table*}[h]
    \centering
    \resizebox{1\textwidth}{!}{
      \begin{tabular}{l | r | c | c}
        \toprule
         Named Entity Category & \# Entities & Belongs to & Examples \\
        \midrule
        Actor & 660 & Celebrity & Sean Connery, Kim Hyun-joong\\
        Restaurant & 237 & Business Brand & Panda Express, KFC\\
        Church & 215 & Landmark & Wolvendaal Church, Innvik Church \\
        Television actor & 157 & Celebrity & Simon Cowell, Whitney Port \\
        Politician & 149 & Celebrity & Boris Johnson, Barack Obama \\
        Singer & 146 & Celebrity & Seun Kuti, Shreya Ghoshal\\
        Football Player & 143 & Celebrity & Marco Reus, James Milner\\
        Bank & 130 & Business Brand & DBS Bank, Lloyds Bank\\
        Airline & 130 &  Business Brand & Air Tahiti, Zambezi Airlines \\
        Supermarket & 128 &  Business Brand & Mercadona, Piggly Wiggly \\
        Retail Store & 124 &  Business Brand & Spencer's Retail, Conad\\
        Film Actor & 116 & Celebrity & Paul Rudd, Anil Kapoor\\
        Mountain & 88 & Landmark & Mount Majura, Mount Uhud\\
        Museum & 74 & Landmark & Louvre Museum, Bapu Museum\\
        Apparel Store & 65 & Business Brand & Quiksilver, Zara\\
        Singer-songwriter & 59 & Celebrity & Joey Tempest, Tuomas Holopainen\\
        Lake & 49 & Landmark & Lough Key, Qinghai Lake\\
        Model & 47 & Celebrity & Lily Cole, Tyson Beckford\\
        Mosque & 47 & Landmark & The Fatih Mosque, Ahl Fas Mosque\\
        Castle & 46 & Landmark & Dunsany Castle, Egeskov Castle\\
        Park & 45 & Landmark & Cove Island Park, Baishamen Park \\
        Auto showroom & 38 & Business Brand & Honda, Volkswagen\\
        Petrol Station & 35 & Business Brand & Petrobras, Petro-Canada\\
        Comedian & 34 & Celebrity & Kapil Sharma, Ken Jeong\\
        Building & 33 & Landmark & De Bazel, ASEM Tower\\
        \bottomrule
      \end{tabular}}
      \caption{\label{tab:entity_type_statistics} Distribution of the top 25 most frequent categories of named entities present in the \data{} dataset.}
\end{table*}

\begin{figure*}[!t]
\centering
\includegraphics[width=1.0\textwidth]{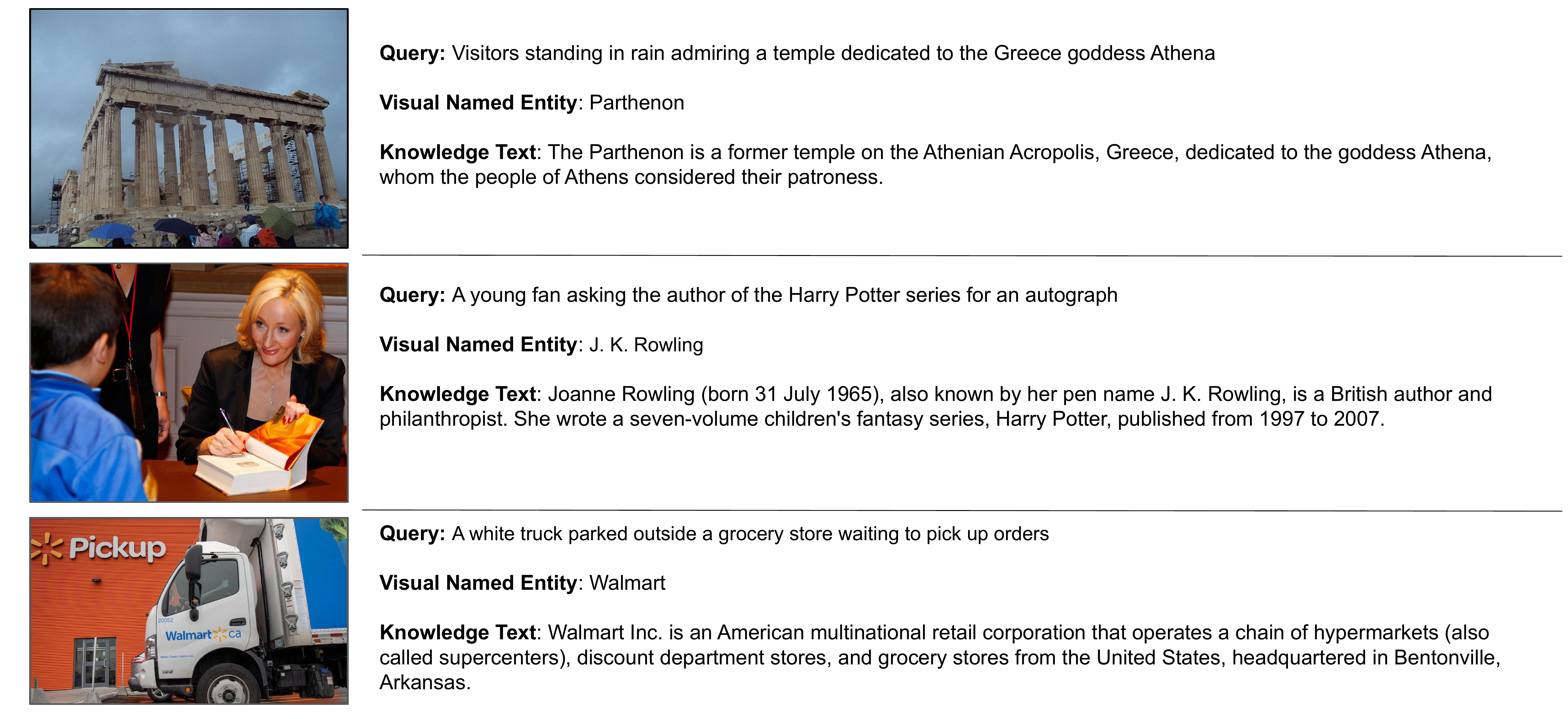}
\caption{\label{fig:example-query-knowledge} \textbf{A selection of examples from \data{}} along with the ground truth visual named entities present in the images and the associated knowledge texts extracted from their respective Wikipedia articles.}
\end{figure*}

\begin{table*}[t]
    \centering
    \resizebox{0.9\textwidth}{!}{
      \begin{tabular}{c c c c c c }
        \toprule
        Type & Number of & Avg. Length of & Avg. Length of & Number of & Number of \\
        & Named Entities & Knowledge (words) & Queries (words) & Countries & Entity types \\
        \midrule
        Brand & 1060 & 44.2 & 11.7 & 79 & 39 \\
        Celeb & 2000 & 39.0 & 14.0 & 92 & 150 \\
        Landmark & 2000 & 41.7 & 13.6 & 40 & 463 \\
        \bottomrule
      \end{tabular}}
      \caption{\label{tab:cofar_type_statistics} Statistics about the three categories of data in \data{}.}
\end{table*}

\begin{table*}[t]
    \centering
    \resizebox{0.7\textwidth}{!}{
      \begin{tabular}{l c c c c c c c c }
      \toprule
        \multicolumn{1}{c}{} & \multicolumn{4}{c}{COFAR-1K (Unseen entities)} & \multicolumn{4}{c}{COFAR-1K (Seen entities)} \\
        \cmidrule(r){2-5}
        \cmidrule(r){6-9}
        Method & R1 & R5 & R10 & MdR & R1 & R5 & R10 & MdR \\
        \midrule
        \model{} & 31.6 & 64.4 & 76.2 & 3 & \textbf{35.1} & \textbf{72.6} & \textbf{88.6} & \textbf{3} \\
        \bottomrule
      \end{tabular}}
      \caption{\label{tab:seen_vs_unseen_cofar} Performance of \model{} on two COFAR-1K versions comprising of entities previously unseen during training and entities seen during training. We observe that performance of \model{} is higher for already-seen entities.}
\end{table*}

\begin{figure*}[h]
\centering
\includegraphics[width=0.9\textwidth]{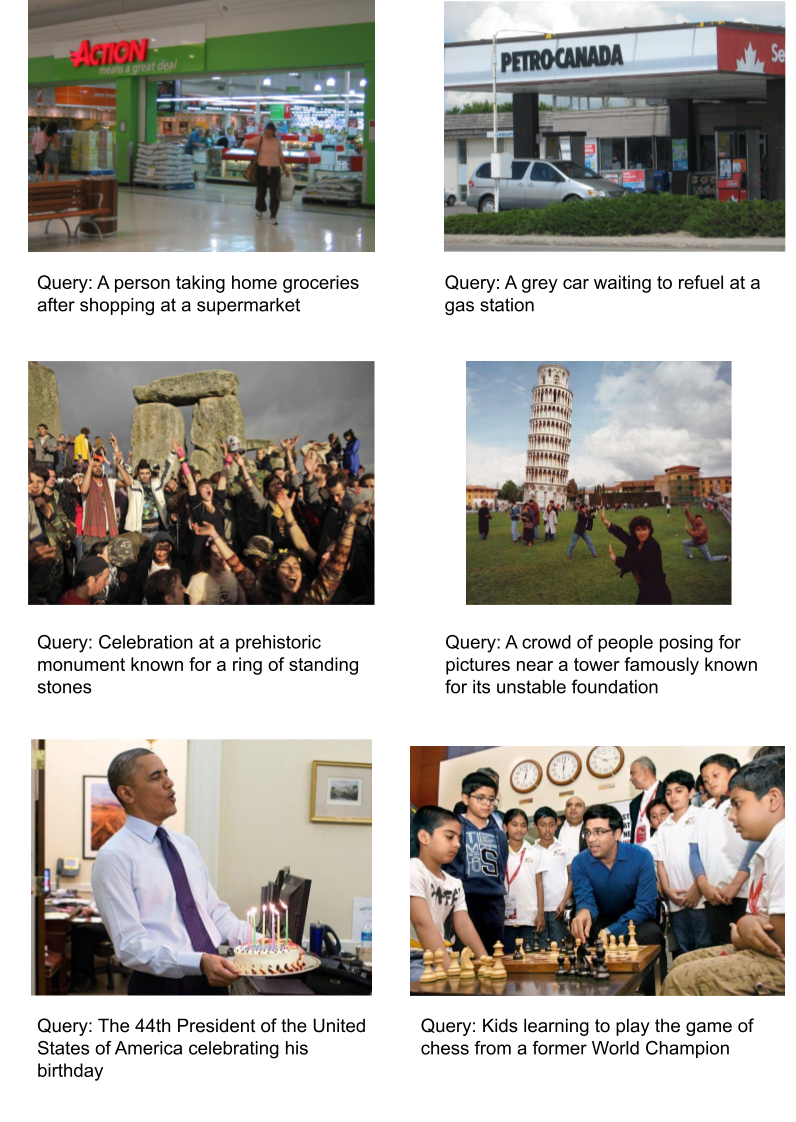} 
\caption{\label{fig:dataSample} \textbf{A selection examples from \data{}}}
\end{figure*}

\end{document}